\documentclass[pdflatex,./bst/sn-mathphys-num]{sn-jnl}

\usepackage{graphicx}%
\usepackage{multirow}%
\usepackage{amsmath,amssymb,amsfonts}%
\usepackage{amsthm}%
\usepackage{mathrsfs}%
\usepackage[title]{appendix}%
\usepackage{xcolor}%
\usepackage{textcomp}%
\usepackage{manyfoot}%
\usepackage{booktabs}%
\usepackage{algorithm}%
\usepackage{algorithmicx}%
\usepackage{algpseudocode}%
\usepackage{listings}%
\usepackage{natbib}
\usepackage{subcaption}
\usepackage{url}

\theoremstyle{thmstyleone}%
\theoremstyle{thmstyletwo}%

\theoremstyle{thmstylethree}%

\begin{document}

\title[Article Title]{Deep Sub-Ensembles Meets Quantile Regression: Uncertainty-aware Imputation for Time Series}

\author[1]{\fnm{Ying} \sur{Liu}}\email{liuying1@mail.hfut.edu.cn}

\author*[2]{\fnm{Peng} \sur{Cui}}\email{cui-peng@mail.tsinghua.edu.cn}

\author*[1]{\fnm{Wenbo} \sur{Hu}}\email{wenbohu@hfut.edu.cn}

\author[1]{\fnm{Richang} \sur{Hong}}\email{hongrc@hfut.edu.cn}

\affil[1]{\orgdiv{Department of Computer and Information}, \orgname{Hefei University of Technology}, \orgaddress{\street{485 Danxia Rd}, \city{Shushan Qu}, \postcode{230601}, \state{Heifei Shi}, \country{China}}}

\affil[2]{\orgdiv{Dept. of Comp. Sci. \& Tech.}, \orgname{Tsinghua University}, \orgaddress{\street{30 Shuangqing Rd}, \city{Haidian Qu}, \postcode{100084}, \state{Beijing Shi}, \country{China}}}


\abstract{Real-world time series data often exhibits substantial missing values, posing challenges for advanced analysis. A common approach to addressing this issue is imputation, where the primary challenge lies in determining the appropriate values to fill in. While previous deep learning methods have proven effective for time series imputation, they often produce overconfident imputations, which poses a potentially overlooked risk to the reliability of the intelligent system. Diffusion methods are proficient in estimating probability distributions but face challenges under a high missing rate and are, moreover, computationally expensive due to the nature of the generative model framework. In this paper, we propose Quantile Sub-Ensembles, a novel method that estimates uncertainty with ensembles of quantile-regression-based task networks and incorporate Quantile Sub-Ensembles into a non-generative time series imputation method. Our method not only produces accurate and reliable imputations, but also remains computationally efficient due to its non-generative framework. We conduct extensive experiments on five real-world datasets, and the results demonstrates superior performance in both deterministic and probabilistic imputation compared to baselines across most experimental settings. The code is available at \url{https://github.com/yingliu-coder/QSE}.}

\keywords{Time series imputation, Uncertainty quantification, Deep ensembles, Quantile regression}



\maketitle

\section{Introduction}\label{sec1}

Multivariate time series are used in a variety of real-world applications, including meteorology~\citep{weather2}, financial marketing~\citep{financial2}, material science~\citep{gao2024dilated}, and artificial intelligence for IT Operations (AIOps)~\citep{li2022stackvae}. Time series data serves as prevalent signals for classification and regression tasks across various applications. However, missing values commonly appear in time series due to data corruptions, merging irregularly sampled data, and human operational errors. This has been recognized as a severe problem in time series analysis and downstream applications. 

The traditional time series imputation methods are divided into two categories. 
One is deletion, which removes partially observed samples and features, resulting in biased parameter estimates. The other is the statistical method, where missing values are imputed using median or average values. However, this approach is unsuitable for high-precision scenarios. Consequently, numerous studies have resorted to machine-learning approaches to address the assignment of imputing missing values. However, some studies show that traditional machine-learning methods such as K-Nearest Neighbor (KNN)~\citep{KNN}, Multivariate Imputation by Chained Equations (MICE)~\citep{MICE}, and Matrix Decomposition~\citep{matricdec} are not optimal for large-scale data.

In recent years, deep neural networks have achieved remarkable success in time series imputation. RNN-based models can achieve high accuracy by capturing temporal dependencies of time series in imputing missing values~\citep{BRITS, TKAE}. Nevertheless, they cannot quantify uncertainty and derive a reliable imputation rule. To address this issue, Bayesian Neural Networks (BNNs) estimate uncertainty in their parameter space by integrating Bayesian probability theory to impute missing values~\citep{BNNI}. However, performing Bayesian inference over the high-dimensional network weights is computationally expensive and challenging. Recently, the score-based diffusion model such as CSDI~\citep{CSDI} has been proposed to learn the conditional distribution for probabilistic imputation.
Diffusion models primarily rely on the generative distribution learned from the observations, which may lead to degenerative results when handling data with a high missing rate~\citep{cao2024survey}. In addition, the generative diffusion model entails sampling-intensive Markov chain iterations which fundamentally contribute to the prohibitive computational costs during the inference process.

In view of this, we introduce Quantile Sub-Ensembles, which produces high-quality uncertainty estimates with ensembling quantile-regression-based task networks of the model, while sharing a common trunk network. It offers a robust solution to the phenomenon of overfitting to the mean and extreme values. Subsequently, we equip bidirectional long short-term memory networks (BiLSTM) with Quantile Sub-Ensembles to quantify inherent uncertainty while making accurate imputation. Empirically, we conduct extensive experiments on five real-world datasets (solar, traffic, healthcare, electricity, and air-quality) to evaluate the performance of the proposed method, which shows that our method outperforms the baseline models on the most evaluation metrics in terms of imputation accuracy and uncertainty quantification. Moreover, the proposed method delivers substantial computational advantages with significantly faster training and inference speeds compared to generative diffusion models.

The main contributions are summarized as follows: 
\begin{itemize}
\item[$\bullet$] We develop Quantile Sub-Ensembles, an innovative method for estimating uncertainty through an ensemble of quantile-regression-based task networks, which effectively avoids overfitting to sample means and extreme values.

\item[$\bullet$] We incorporate Quantile Sub-Ensembles into BiLSTM for probabilistic time series imputation, which can estimate the predictive uncertainty effectively during the imputation process.

\item[$\bullet$] We evaluate the proposed method on five representative datasets, which shows our method not only achieves high imputation accuracy but also produces reliable uncertainty especially in a higher missing rate. Moreover, it is significantly more computationally efficient than diffusion models.

\end{itemize}

\section{Related works}

\subsection{Statistical methods}
Over the past few decades, numerous imputation methods have been employed to address missing values in multivariate time series. Statistical methods are widely used for time series imputation due to their user-friendly nature and computational efficiency. For example, since there is a strong tendency for time series, we can impute missing values through the trend of the fitted data changes, which is called linear interpolation~\citep{linearInterpolation}. Because of their simplicity, these imputation methods are often inadequate in situations that demand high precision. Autoregressive methods such as ARIMA~\citep{ARIMA} are applied to fill the missing values. It converts the data into stationary data through differential analysis, and then regresses the dependent variable only on its lagged value, as well as the present value and lagged value of the random error term. These methods often struggle to effectively impute missing values due to their neglect in capturing dependencies across time points and the relationship between observed and imputed values. 

\subsection{Traditional machine-learning-based approaches}
Traditional machine-learning-based approaches are widely applied to time series imputation. For instance, linear regression models impute missing values by leveraging the linear relationship between the target variable and one or more input features.
Lagged values or relevant features can be used as inputs for imputation. The K-Nearest Neighbors (KNN) algorithm~\citep{KNN} can also be applied to the time series imputation such as MedImpute~\citep{bertsimas2021imputation}, which imputes missing data by integrating patient-specific temporal information into a flexible framework. While KNN is a straightforward and interpretable imputation method for time series data, it has limitations related to parameter sensitivity, computational complexity and its ability to handle various data characteristics effectively. Multivariate Imputation by Chained Equations (MICE)~\citep{MICE} initializes the missing values arbitrarily and each missing variable is estimated according to the chain equations, which may not be suitable for large-scale data. 

\subsection{Deep learning models}
Recently, deep learning models have demonstrated powerful learning capabilities on multivariate time series imputation. Recurrent Neural Networks (RNNs) have been optimized to capture the temporal dependencies and do not impose specific assumptions. For example, BRITS~\citep{BRITS} imputes missing values according to hidden states from BRNNs, which learns the temporal dependencies from time series directly.
TKAE~\citep{TKAE} generates compressed representations of multivariate time series through a recurrent neural network-based autoencoder to effectively handle missing data. The self-attention-based model, SAITS~\citep{SAITS}, learns missing values by a joint-optimization training approach of imputation and reconstruction. Moreover, the self-attention mechanism is incorporated into RNNs~\citep{GLIMA} to capture global and local information from time series and distant dependencies. While these methods have high accuracy, this is not sufficient to characterize the performance in time series imputation where uncertainty factors can influence the performance tremendously.

Bayesian Neural Networks (BNNs) are particularly well-suited for uncertainty estimation~\citep{BNNI}. Unlike traditional neural networks, BNNs provide probabilistic distributions over their weights and biases, making them valuable for tasks that require uncertainty quantification and robust predictions. However, an incorrect prior choice may lead to poor model performance or difficulty in convergence~\citep{bayesianprior}.
Denoising diffusion probabilistic models~\citep{Diffusion,CSDI,Score-CDM} have surpassed existing models in many deep learning tasks. As an innovative imputation approach, they offer probabilistic estimation and learn the conditional distribution using conditional score-based diffusion models. However, there are instances when their benefits can also translate into their drawbacks. Diffusion models for imputation may face challenges when handling data with a high missing rate, as the generative model lacks sufficient data points to infer the distributional characteristics of missing values~\citep{cao2024survey}.
In addition, the high computational complexity of the probabilistic inference process in conditional score-based diffusion models leads to slow training convergence and substantial consumption of computing resources.

\subsection{Quantile regression}

Quantile regression techniques~\citep{QR-DQN} have been introduced to learn the probability distribution of state-action values, facilitating the modeling of risks across various confidence intervals and fortifying resilience against extreme scenarios. This approach, shared across the referenced works, revolves around the pivotal step of apportioning layers among different quantiles, a methodology also embraced within the framework of this study. Quantile regression is less sensitive to outliers compared to ordinary least squares (OLS) regression~\citep{yadav2024exponential}. It provides a more reliable estimation by using the conditional quantiles compared to OLS, making it suitable for data with extreme values or skewness characteristics. In traditional quantile regression, the model assumes constant variance. However, in our model, we observe that the variance under different quantiles varies with the input variables. This leads us to adopt a heteroscedasticity assumption, which allows for a more flexible and accurate modeling of the data~\citep{Heteroscedasticity}. Based on these advantages, quantile regression is widely used in economics and finance to analyze income distribution~\citep{income}, asset pricing~\citep{pricing} and risk assessment~\citep{risk}. By employing quantile regression, our model accounts for the initial distribution of the data when filling in missing values. This narrows the range of imputed values, reducing their dispersion. This method effectively tackles the inherent uncertainty present in the data. In this scenario, each quantile represents a specific percentile, such as the median (50th percentile) or the 90th percentile. The width of the estimated quantile (i.e., prediction intervals) can provide valuable information about the uncertainty associated with the predictions~\citep{calibration}.

\section{Methodology}
In this section, we first establish the theoretical foundation by analyzing uncertainty quantification mechanisms in Deep Ensembles and their parameter-efficient variant Deep Sub-Ensembles. Subsequently, we present the integration of quantile regression with Deep Sub-Ensembles, yielding the proposed Quantile Sub-Ensembles that generates uncertainty estimates through parallel task networks. We then implement this approach by enhancing the BiLSTM architecture with Quantile Sub-Ensembles for uncertainty-aware time series imputation. Finally, we provide theoretical discussions for this method.

\begin{figure*}[htbp]
    \centering
    \begin{subfigure}[b]{0.45\textwidth}
        \centering
        \includegraphics[width=0.8\linewidth]{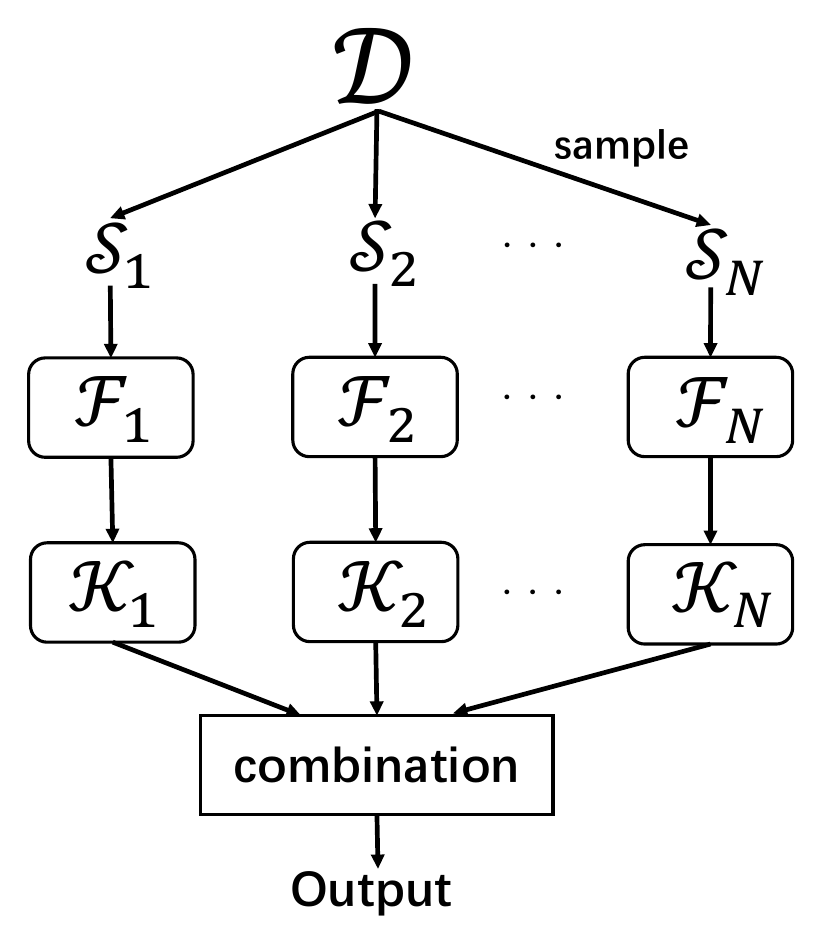}
        \caption{Deep Ensembles}
        \label{fig:Deep Ensembles}
    \end{subfigure}
    \hspace{0.02\textwidth}
    \begin{subfigure}[b]{0.45\textwidth}
            \centering
        \includegraphics[width=0.9\linewidth]{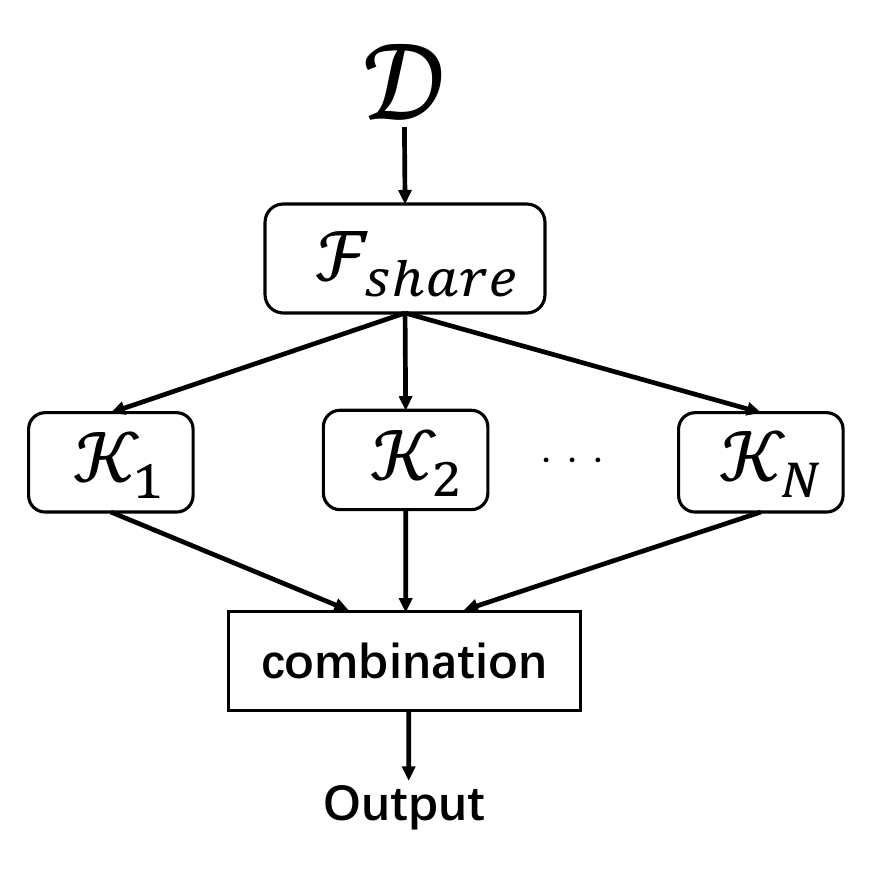}
        \caption{Deep Sub-Ensembles}
        \label{fig:Deep Sub-Ensembles}
    \end{subfigure}
    \caption{A conceptual comparison between Deep Ensembles and Deep Sub-Ensembles with $N$ ensemble members is illustrated. In (a), multiple trunk networks, denoted as $\mathcal{F}_i$ are independently trained on sample data $\mathcal{S}_i$ from dataset $\mathcal{D}$. In contrast, (b) shows that a single trunk network, $\mathcal{F}_{share}$, is shared across all ensemble members. $\mathcal{K}_i$ represent task networks with $n$ ensemble members. Both methods involve combining ensemble predictions to generate outputs that reflect uncertainty.}
    \label{fig:Ensembles}
\end{figure*}

\subsection{Deep Ensembles and Deep Sub-Ensembles}
Deep Ensembles~\citep{deepEnsembles} is a straightforward yet highly effective approach for quantifying predictive uncertainty. Deep Ensembles consists of multiple trunk networks that are trained independently on sample data and combines ensemble predictions by uniform voting which is shown in Fig.~\ref{fig:Deep Ensembles}. Deep Ensembles is widely used to improve model robustness and performance~\citep{de1} by alleviating the over-fitting of DNNs and capturing different patterns in the data, ultimately enhancing the reliability and accuracy of predictions. Performing Bayesian inference over DNNs weights (i.e., learning weight distributions) is very challenging in highly nonlinear and high-dimensional space of DNNs. Deep Ensembles directly combines the outputs of different DNNs and are a prevalent tool for producing uncertainty, which is well-suited for distributed computing environments and makes it particularly appealing for large-scale deep learning applications owing to the simple implementation. Besides, Deep Ensembles offer a valuable means of estimating predictive uncertainty stemming from neural networks and effectively calibrating unidentified classes within supervised learning problems~\citep{deepEnsembles}. 

Due to the parallel training and inference across multiple neural network models with the same structure in Deep Ensembles, the computational overhead becomes enormous as ensemble members increase. Deep Sub-Ensembles~\citep{sub-ensembles} is motivated by the observation that deep neural networks inherently learn a hierarchical structure of features, where the complexity of features increases with depth, culminating in the final layers that execute specific tasks like classification or regression. In this case, a neural network can be separated into two sub-networks, the trunk network $\mathcal{F}_{share}$ which tends to learn similar features across different ensemble members, and the task network $\mathcal{K}$ which play a more significant role in contributing to uncertainty.
The objective of Deep Sub-Ensembles is to create an ensemble comprising a shared trunk network $\mathcal{F}_{share}$ and multiple instances of task-specific networks $\mathcal{K}_i$ to form a sub-ensemble. The architecture is shown in Fig.~\ref{fig:Deep Sub-Ensembles}. In this paper, we adopt Deep Sub-Ensembles, which only ensembles the task network while sharing the trunk network. This method  still produces high quality uncertainty estimates, yet significantly accelerates inference, since the trunk network requires only one forward pass and the smaller task network requires multiple but lightweight passes~\citep{sub-ensembles}. Consequently, the model efficiency is improved.

\subsection{Quantile Sub-Ensembles}
\begin{figure}[h]
    \centering
    \includegraphics[width=0.9\textwidth]{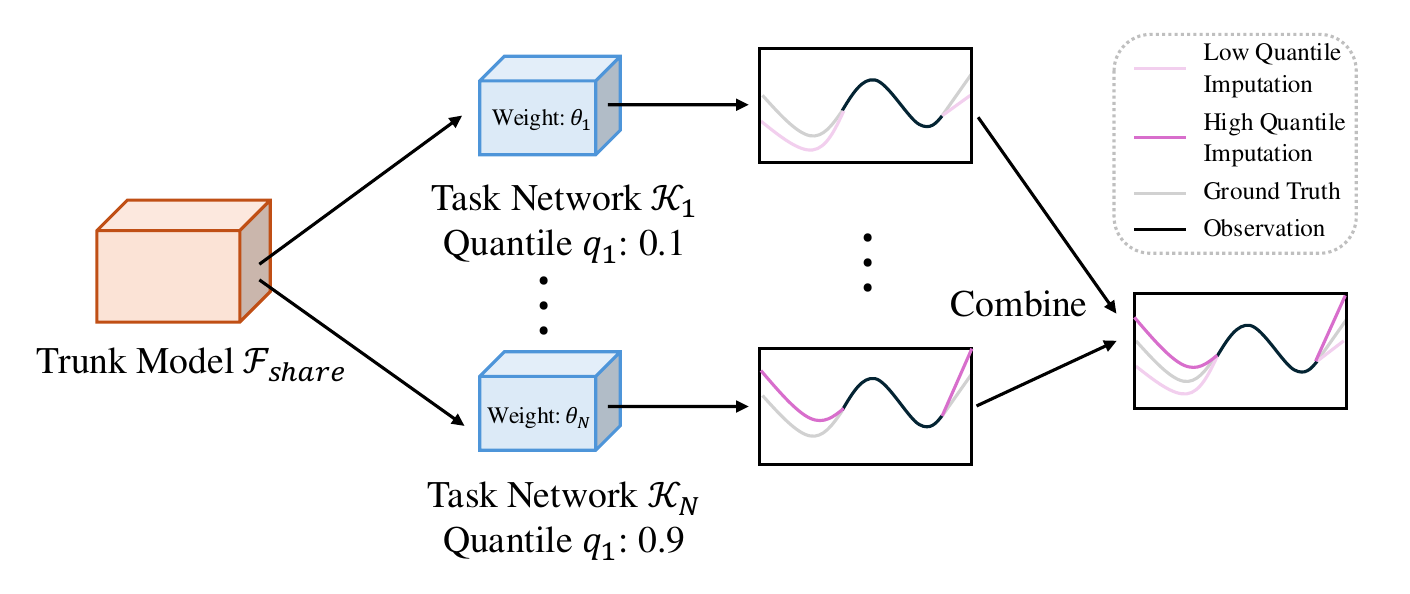}
    \caption{The framework of Quantile Sub-Ensembles. We stack the trunk network $\mathcal{F}_{share}$ and $N$ parallel task networks $\{\mathcal{K}_1, \mathcal{K}_2, ..., \mathcal{K}_N \}$ with the same architecture corresponding to $N$ ensemble members. The trunk network $\mathcal{F}_{share}$ functions as a shared feature extractor that learns temporal patterns from the input time series. This design not only reduces redundant parameterization but also improves efficiency. Subsequently, the task networks learn quantiles $\{ q_1, q_2,..., q_N\}$ through randomly initialized parameters $\{\theta_1, \theta_2,...,\theta_N\}$. The imputations of all quantiles will be combined to obtain the reliable result.}
    \label{fig:Quantile Sub-Ensembles}
\end{figure}

In time series, values can be missing for either continuous segments or sporadic time points. Such random missing patterns introduce significant uncertainty into our imputation results, which can adversely affect the performance of downstream tasks.
We develop Quantile Sub-Ensembles, an algorithm to quantify uncertainty with training multiple quantile-regression-based task networks and alleviate overfitting to the mean sample intensity and extreme values. The concept is shown in Fig.~\ref{fig:Quantile Sub-Ensembles}. 

Quantile regression~\citep{QR} is a statistical technique that extends traditional linear regression by modeling different quantiles of the response variable's distribution. While linear regression focuses solely on estimating the conditional mean, quantile regression aims to estimate the conditional quantiles (e.g., the 10th or 90th percentiles). This approach provides a more nuanced understanding of the relationship between predictors and the response variable, offering insights across various points of the distribution rather than just the mean.

Quantile regression aims to estimate a specific quantile of the response variable. The conditional distribution function~\citep{CQR} of $Y$ given $X = x$ is: 
\begin{equation}
    F(y|X) := P(Y \leq y | X = x),
\end{equation}
and the $\alpha$th conditional quantile function is:
\begin{equation}
    Q_{\alpha}(x) := \inf \{ F(y|X) \geq \alpha \}.
\end{equation}
To learn the quantile function and estimate the corresponding values with various quantiles. Quantile regression loss is applied:

\begin{equation}
    l(x, y, q_i) = 
    \begin{cases}
        q_i (x - y) & if \quad x \geq y \\
        (1 - q_i) (y - x) & if \quad x < y
    \end{cases},
\end{equation}
where $q_i \in \{ q_1, q_2,...,q_N \}$ is a set of quantiles which has already been preconfigured and will not be optimized. 

Once the quantile regression loss for each member is calculated, the parameters $\theta_1, \theta_2, ... , \theta_N$ are updated through backpropagation and gradient descent. The gradients computed from the quantile regression loss $l_i$ help in adjusting the parameters $\theta_i$ of each task network in a way that the imputations align more closely with the quantile $q_i$. By training each ensemble member to minimize its corresponding quantile regression loss, the structure ensures that different task network focus on different quantiles of the predictive distribution. 

We consider the ensemble as a mixture model~\citep{deepEnsembles, mikalsen2018time} with uniform weighting across its components and combine the outputs as $p(y|x) = N^{-1} \sum_{i=1}^N p_{\theta_i}(y|x, \theta_i)$. For regression tasks, the output is modeled as a mixture of Gaussian model with quantiles are derived from the cumulative distribution of a single fitted distribution. Therefore, our method naturally guarantees ordered quantiles without the risk of crossing. To simplify the computation of quantiles and predictive probabilities, we approximate the ensemble output with a Gaussian distribution whose mean $u_{\theta_i}(x)$ and variance $\sigma_{\theta_i}^2(x)$ correspond to the mean $u_*(x)$ and variance $\sigma_*^2(x)$ of the mixture:
\begin{equation}
    fp(y|x) \sim \mathcal{N}(u_*(x), \sigma_*^2(x)),
\end{equation}
\begin{equation}
    u_*(x) = N^{-1} \sum_i u_{\theta_i}(x),
\end{equation}
\begin{equation}
    \sigma_*^2(x) = N^{-1} \sum_i (\sigma_{\theta_i}^2(x) + u_{\theta_i}^2(x)) - u_*^2(x).
\end{equation}
Regression necessitates a distinct loss function, as supervision is available for $u_*(x)$, but not for $\sigma_*^2(x)$. To address this, we integrate a heteroscedastic Gaussian log-likelihood loss~\citep{uncertaintiesneed, deepEnsembles} into quantile regression loss:
\begin{equation}
    -logp(y|x) = \frac{log \sigma_*^2(x)}{2} + \frac{N^{-1}\sum_{i=1}^{N}{l_i(u_{\theta_i}(x), y, q_i)}}{2\sigma_*^2(x)}.
\end{equation}
The quantile regression loss $l_i(u_{\theta_i}(x), y, q_i)$ is defined as:
\begin{equation}
    l(u_{\theta_i}(x), y, q_i) = \left(q_i * |\max(u_{\theta_i}(x) - y, 0)| + (1 - q_i) * |\max(0, y - u_{\theta_i}(x))|\right).
\end{equation}
With this loss function, the output variance $\sigma_*^2(x)$ represents the predictive uncertainty estimate.

\subsection{Bidirectional Recurrent Imputation with Quantile Sub-Ensembles}

\begin{figure}[ht]
    \centering
    \includegraphics[width=\textwidth]{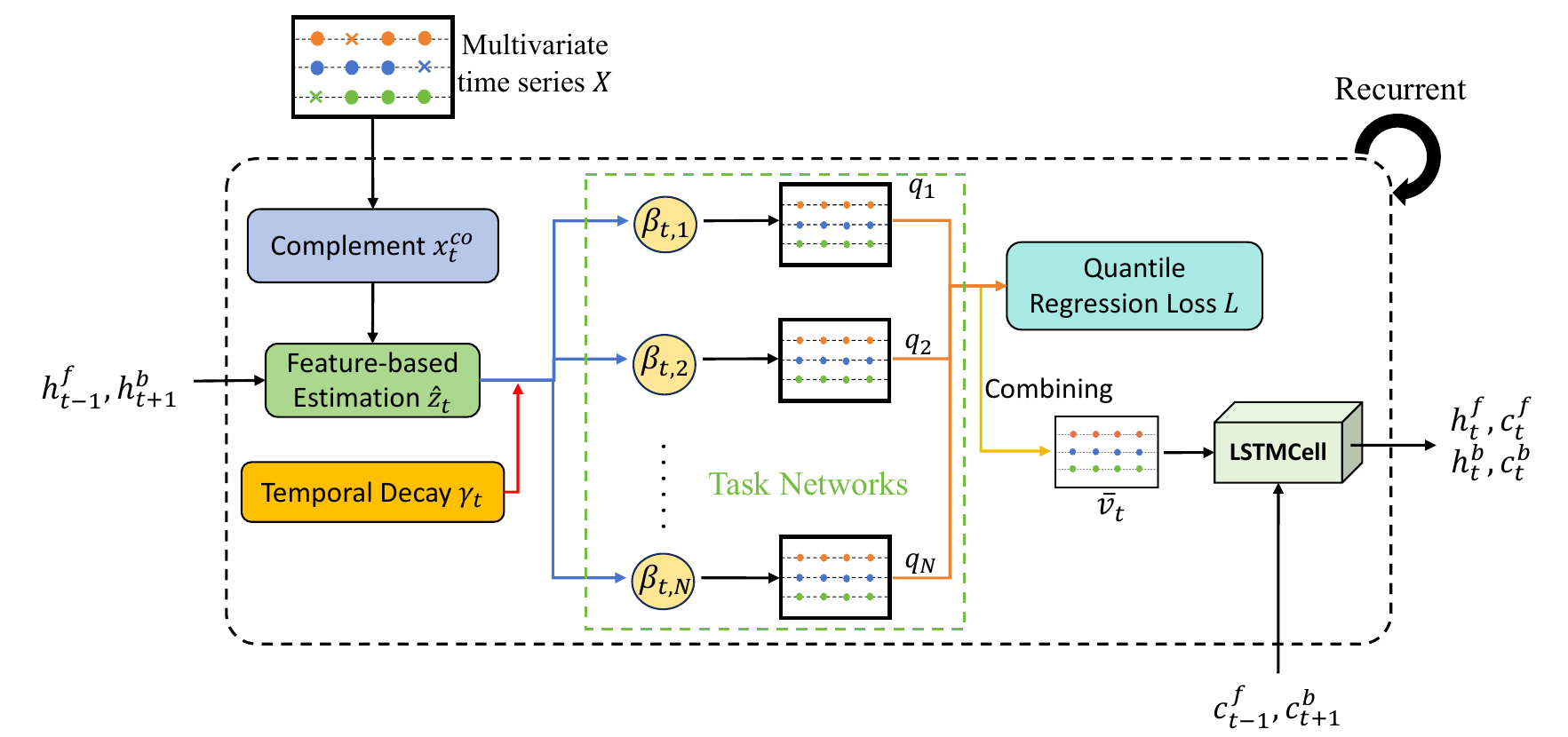}
    \caption{The overview of the model framework. The multivariate time series $X$ with missing values are fed to the complement layer which fills missing values. Next, the output of the complement layer $x_t^{co}$ and the previous hidden state $h_{t-1}^f, h_{t+1}^b$ is delivered to the feature-based estimation layer to capture the feature relationship. Then we combine $\hat{z}_t$ and the temporal decay factor $\gamma_t$ which gradually diminishes history information over time as the input to $N$ task networks corresponding to $N$ ensemble members. 
    In the meanwhile, quantile regression loss $L$ is computed to update the parameters through the process of backpropagation. Finally, the imputed time series of each ensemble is combined and fed into BiLSTM with cell state $c_{t-1}^f, c_{t+1}^b$.}
    \label{fig:framework}
\end{figure}

In this section, we introduce a Bidirectional Long Short-Term Memory network (BiLSTM) enhanced with Quantile Sub-Ensembles for time series imputation. The BiLSTM processes input data in both forward and backward directions simultaneously, allowing it to capture contextual information from both past and future data points. By incorporating Quantile Sub-Ensembles, the model estimates multiple quantiles that capture uncertainty in the data, leading to more reliable imputation performance and reducing the risk of overfitting in complex time series scenarios. The overall architecture of the model is detailed in Fig.~\ref{fig:framework}.

We denote the input multivariate time series
as $X = \{x_{1:T}^{1:K} \} \in 
\mathbb{R}^{T \times K}$ where $T$ is the length of the time series, $K$ is the feature dimension, and $x_t^k$ represents the $k$-th feature value at the timestamp $d_t$. To represent the missing values in $X$, we define a mask matrix $M = \{m_{1:T}^{1:K} \} \in \mathbb{R}^{T \times K}$, where $m_t^k = 1$ if $ x_t^k $ is observed and $m_t^k = 0$ if $x_t^k$ is missing. In the time series imputation task, our target is to fit an imputation function $F$: $X \longrightarrow \hat{X}$, where $\hat{X} = \{\hat{x}_{1:T}^{1:K}\} \in \mathbb{R}^{T \times K}$ is the output imputed time series.

In some instances, certain variables may be absent in consecutive timestamps. Therefore, the time gap from the last observation to the current timestamp $d_t$ can be denoted as:


$$
\delta_t^k = \begin{cases} 0 , & \text{if }  t = 1, \\ d_t - d_{t-1} + \delta_{t-1}^k, & \text{if }  t > 1  \text{ and }  m_{t-1}^k = 0,  \\ d_t - d_{t-1}, & \text{if }  t > 1  \text{ and }  m_{t-1}^k = 1.
\end{cases}
$$

In this paper, we focus on probabilistic time series imputation~\citep{GPVAE}, i.e., estimating the probabilistic distribution of time series missing values through observation. Therefore, uncertainty can be quantified while imputing missing values.

In our case, on account of missing values in time series $X$, we can't denote $X_t$ directly as the input to the model. The solution to this problem is to use a ``complement" input. In this initial step, we don't need to focus too much on the accuracy of the imputed values. It's sufficient to use a linear layer:
\begin{eqnarray}
    x_t' &=& W_x h_{t-1} + b_x, \\
    x_t^{co} &=& m_t \odot x_t + (1 - m_t) \odot x_t',
\end{eqnarray}
where $x_t^{co}$ is the ``complement" time series data   and it's dimension is $n \times d$ ($n$ is the size of batch and d is the dimension of features). 

It has been observed that if variables are missing for a substantial period of time,  the influence of the input variables will fade away over time. Therefore, we introduce a \textit{temporal decay factor} $\gamma_t$~\citep{2018Recurrent} to consider the following important factors:
\begin{equation}
    \gamma_t = \exp\{ -\max(0, W_{\gamma} \delta_t) + b_{\gamma}\},
\end{equation}
where $W_{\gamma}, b_{\gamma}$ is trainable weight parameters.

In order to utilize feature dimension information during imputation, we introduce feature-based estimation $\hat{z}_t$ ~\citep{BRITS}:
\begin{equation}
    \hat{z}_t = W_z x_t^{co} + b_z,
\end{equation}
where $W_z, b_z$ are trainable weight parameters. 

Based on it, we can combine the historical-based estimation $x_t'$ and the feature-based estimation $\hat{z}_t$. Then we define $\hat{v}_{t,1}, \hat{v}_{t,2},..., \hat{v}_{t,N}$ as the vectors of task networks:
\begin{eqnarray}
    \beta_{t,i} = \sigma(W_{\beta,i}[\gamma_t^{\beta} \circ m_t] + b_{\beta, i}), \\
    \hat{v}_{t,i} = \beta_{t,i} \odot \hat{z}_t + (1 - \beta_{t,i}) \odot x_t',
\end{eqnarray}
where $\beta_{t,1}, \beta_{t,2},..., \beta_{t,N}$ is the weights. The function of the task networks layers (12) and (13) is to combine the historical-based estimation $x_t'$ and the feature-based estimation $\hat{z}_t$. Adding some layers between (12) and (13) will not promote the information combination, but increase the computational overhead. We can replace missing values in $x_t$ with the corresponding values in $\hat{v}_{t,i}$:
\begin{equation}
    v_{t,i} = m_t \odot x_t + (1-m_t) \odot \hat{v}_{t,i},
\end{equation}
where $v_{t,1}, v_{t,2},..., v_{t,N}$ represent the imputed values produced by models with the same structure but different parameters. Therefore, we can leverage them to compute the quantile regression loss:
\begin{equation}
    \resizebox{.91\linewidth}{!}{$
            L(q_i, x_t, m_t, v_{t,i}) = \sum_{i = 1}^{N} \sum_{t = 1}^{T}  (q_i * |\max(x_t - v_{t,i}, 0)| + (1 - q_i) * |\max(0, v_{t,i} - x_t)|) \odot m_t
        $}.
\end{equation}

Then we feed $\overline{v_{t}}$, the mean of $v_{t,1}, v_{t,2},..., v_{t,N}$, to the next bidirectional recurrent long short-term memory networks(BiLSTM) to learn the mixed information of all features and capture the short and long relationships of different variables. The overall procedure of our algorithm is summarized in Algorithm~\ref{algorithm_2}.

\subsection{Theoretical Discussions}

\textbf{Limitation of quantiles.} The Bayesian Gaussian Mixture model employed for calculating quantiles utilizes a maximum likelihood estimation on the posterior distribution, thereby achieving convergence to the desired $N$ quantiles. However, increasing the number of quantiles corresponds to a greater overlap between the intensity distributions of adjacent quantile intervals which can potentially lead to overfitting. This occurs because the model becomes more complex with more quantiles, potentially fitting noise in the data rather than capturing the underlying relationship~\citep{fakoor2023flexible}. As the number of quantiles increases, the model's flexibility grows, which can lead to higher variance and poorer generalization.

\begin{algorithm}[h]
  \caption{The overall forward procedure of our algorithm}
  \label{alg::conjugateGradient}
  
  \begin{algorithmic}[2]
    \Require
       Multivariate time series $X = \{x_{1:T}^{1:K} \} \in 
    \mathbb{R}^{T \times K}$, number of ensmeble members $N$,  quantiles $\{q_1, q_2,..., q_N \}$, iterations $T$.
    \Ensure
      Imputed multivariate time series $V = \{\overline{v}_{1:T}^{1:K} \} \in \mathbb{R}^{T \times K}$.
      
    \State Initialize the parameters $h, c$ to all zeros;   
 
    \For {$t = 1:T$}          
        \State $x_t' = W_x h_{t-1} + b_x$
        \State $x_t^{co} = m_t \odot x_t + (1 - m_t) \odot x_t'$ \Comment{ Fill in the missing values beforehand}
        \State Compute mean absolute error loss: $L_1(x_t^{co}, x_t, m_t)$
        \State Introduce a temporal decay factor $\gamma_t = \exp\{ -\max(0, W_{\gamma} \delta_t) + b_{\gamma}\}$
        \State Introduce feature-based estimation $\hat{z}_t = W_z x_t^{co} + b_z$
        \State Compute mean absolute error loss: $L_2(\hat{z}_t, x_t, m_t)$
        \For{$i = 1:N$}   \Comment{Train independently in parallel}
            \State Compute the combining weight $\beta_{t,i} = \sigma(W_{\beta,i}[\gamma_t^{\beta} \circ m_t] + b_{\beta, i})$
            \State Define combined vectors $\hat{v}_{t,i} = \beta_{t,i} \odot \hat{z}_t + (1 - \beta_{t,i}) \odot x_t'$
            \State $ v_{t,i} = m_t \odot x_t + (1-m_t) \odot \hat{v}_{t,i}$ \Comment{Replace missing values} 

        \EndFor
        \State $L(q_i, x_t, m_t, v_{t,i}) = \sum_{ 1}^{N} \sum_{1}^{T}  (q_i * |\max(x_t - v_{t,i}, 0)| + (1 - q_i) * |\max(0, v_{t,i} -  $ 
        \State $x_t)|) \odot m_t$
        \State Minimize $L_1(x_t^{co}, x_t, m_t) + L_2(\hat{z}_t, x_t, m_t) + L(q_i, x_t, m_t, v_{t,i})$
        \State Compute $\overline{v_{t}}$, the mean of $v_{t,1}, v_{t,2},..., v_{t,N}$
        , $\hat{x}_t=\overline{v_{t}}$ is the imputed value
        \State 
        $h_t^f = \text{LSTM}_f(\overline{v_{t}}, h_{t-1}^f, c_{t-1}^f), h_t^b = \text{LSTM}_b(\overline{v_{t}}, h_{t+1}^b, c_{t+1}^b)$ 
        \State $h_t = [h_t^f;h_t^b]$ \Comment{LSTMs}
    \EndFor
 
  \end{algorithmic}
  \label{algorithm_2}
\end{algorithm}

\textbf{Robustness with high missing rates.} 
When dealing with time series data with a high missing rate, RNN-based networks can capture temporal dependencies from the non-missing parts of the time series to obtain periodic features. Since real-world time series often exhibit significant periodicity, we can leverage this periodicity to impute large consecutive missing segments accurately~\citep{cyclernn}. Meanwhile, the proposed Quantile Sub-Ensembles method alleviate overfitting on the mean sample intensity and extreme values by fitting each quantile-regression-based task network to the corresponding quantile interval, which provide robust and reliable imputations.

\section{Experiments}
In this section, we compare the proposed method with existing methods in terms of imputation accuracy and uncertainty on time series. We conduct experiments on five public datasets (solar, traffic, healthcare, electricity, and air-quality) from relevant application domains. Subsequently, we also show the hyper-parameters sensitivity and computation efficiency of the proposed method. 

\subsection{Dataset Description}

\textbf{Solar dataset}\footnote{\url{https://www.nrel.gov/grid/solar-power-data.html}} contains power production measurements from 137 photovoltaic (PV) plants located in Alabama, totaling 52,560 timestamps. \textbf{Traffic dataset}~\citep{traffic} comprises the traffic speed of 214 features over a five-day period in Guangzhou, with a sampling rate of one reading per 10 minutes. \textbf{Electricity dataset}\footnote{\url{https://archive.ics.uci.edu/dataset/321/electricityloaddiagrams20112014}} records the hourly electricity consumption (kWh) of 370 clients from January 2011 to December 2014. \textbf{Air-quality dataset}\footnote{\url{https://archive.ics.uci.edu/dataset/501/beijing+multi+site+air+quality+data}}\citep{air-quality} comprises hourly air pollutant observations from 12 monitoring sites in Beijing, spanning March 2013 to February 2017. For each site, 11 continuous time series variables are recorded. \textbf{Healthcare dataset} from \textit{PhysioNet Challenge 2012}~\citep{2013Predicting} consists of 4000 clinical time series with 35 variables. Each time series is recorded within the first 48 hours of the patient's admission to the ICU, so we process it hourly with 48 time steps. Due to the incomplete data collection, solar, air-quality, and healthcare datasets are inherently missing. For example, missing records are often caused by sensor malfunctions. This is likely closer to the MAR (Missing at Random) pattern. The traffic and electricity datasets is complete. For five datasets, we select different missing rates of the available data points as observations under the MCAR (Missing Completely at Random) pattern for model training, and the rest to serve as the ground truth for testing.

\subsection{Experiment Settings}
Following the previous setting~\citep{CSDI}, we use an Adam optimizer with a learning rate of 0.001 and batch size 32 to train our model, which follows previous studies. For all datasets, we normalize the data of all features at time dimension to acquire zero mean and unit variance, which can make model training stable. In addition, we set 5 quantile levels $Q_1 = [0.1, 0.25, 0.5, 0.75, 0.9]$ to calculate the quantile regression loss of each ensemble in parallel.

\subsection{Baselines}
\begin{itemize}
\item[$\bullet$]\textbf{SAITS}~\citep{SAITS} employs a diagonally masked self-attention mechanism to capture temporal dependencies and adopts a joint optimization approach during training.

\item[$\bullet$]\textbf{TimesNet}~\citep{timesnet} addresses the limited representational capacity of 1D time series by transforming them into 2D tensors across multiple periods, thereby extending temporal variation analysis into the 2D space.

\item[$\bullet$]\textbf{Multitask GP}~\citep{2008Multi} learns the covariance between time points and features simultaneously, boosting imputation performance and allowing for flexible modeling of task dependencies.

\item[$\bullet$]\textbf{GP-VAE }~\citep{GPVAE} assumes that a high-dimensional time series can be represented by a low-dimensional Gaussian process that evolves smoothly over time. It reduces non-linear dimensionality and handles missing data with VAE.

\item[$\bullet$]\textbf{V-RIN}~\citep{V-RIN} is a variational-recurrent imputation network that combines an imputation and a prediction network by considering the relevant characteristics, temporal dynamics and uncertainty.

\item[$\bullet$]\textbf{CSDI}~\citep{CSDI} employs score-based diffusion models conditioned on observations. Its probability calculation process can effectively quantify uncertainty and make probabilistic imputation for time series data.

\item[$\bullet$]\textbf{CSBI}~\citep{CSBI} addresses the problem of time series imputation using the Schrödinger bridge problem (SBP), which is gaining popularity in generative modeling. 

\end{itemize}

In addition, we compare the proposed method with the foundational Forward Filling (Forward), Linear Interpolation (Linear), and BiLSTM to demonstrate the superior effectiveness of our method.

\begin{table}[!t]
\renewcommand{\arraystretch}{1.1}
\setlength{\tabcolsep}{2pt}
\caption{Performance comparison of MAE and CRPS at 50\% missing rate.}
    \label{tab:missing_50}
    \centering
    \begin{tabular}{ccc|cc|cc|cc|cc}
\toprule    
  &  \multicolumn{2}{c}{solar}  & \multicolumn{2}{c}{traffic} & \multicolumn{2}{c}{healthcare} & \multicolumn{2}{c}{electricity} & \multicolumn{2}{c}{air-quality} \\ 
  Model  & MAE & CRPS & MAE & CRPS & MAE & CRPS & MAE & CRPS & MAE & CRPS\\
\midrule 
Forward & 2.212 & - & 7.791 & - & 0.923 & - & 1.032 & - & 0.994 & -\\
Linear & 1.975 & - & 5.817 & - & 0.453 & - & 0.734 & - & 0.245 & -\\
SAITS & 1.327  & -  & 3.291  & - & 0.341 & - & 0.876 & - & 0.194 & -\\
TimesNet & 1.221  & - &  3.205 & - & 0.414 & - & 0.305 & - & 0.210 & -\\
BiLSTM & \underline{0.845}  & - &  3.197 & - & 0.348 & - & 0.329 & - & 0.169 & -\\
Multitask GP & 1.706  &  0.203 &  3.532 & 0.092 & 0.516 &  0.581 & 1.066 & 0.913 & 0.313 & 0.342\\
GP-VAE & 1.811  & 0.368  & 3.419  & 0.084 &  0.478 & 0.774 & 1.097 & 0.862 & 0.258 & 0.294\\
V-RIN & 1.547  & 0.362  &  3.375 & 0.107 & 0.365 & 0.831 & 0.541 & 0.583 & 0.197 & 0.268\\
CSDI & 1.104  & 0.166  & 3.202  & 0.076 & \underline{0.315} & \textbf{0.338} & 0.278 & 0.247 & \underline{0.156} & \underline{0.213}\\
CSBI & 1.033  &  \underline{0.153} &  \underline{3.182} & \underline{0.074} & 0.318 & \underline{0.352} & \underline{0.263} & \underline{0.225} & 0.172 & 0.236\\
\textbf{Proposed method}  & \textbf{0.568}  & \textbf{0.085}  &  \textbf{2.884} & \textbf{0.067} & \textbf{0.313} & 0.378 & \textbf{0.212} & \textbf{0.157} & \textbf{0.154} & \textbf{0.191}\\
\bottomrule 
\end{tabular}
\end{table}

\subsection{Experimental Results}
We evaluate imputation performance in terms of mean absolute error (MAE) and continuous ranked probability score (CRPS), and the latter one is regularly adopted to evaluate probabilistic time series forecasting and measure the compatibility of an estimated probability distribution $F$ with an observation $x$. It is
a common and well-defined way to evaluate uncertainty using metrics related to loss functions, for example PICP in~\cite{PICP} which is similar to CRPS in quantifying uncertainty. Thus, CRPS is a reasonable evaluation metric. It can be defined as:
\begin{equation}
    CRPS(F^{-1}, x) = \int_0^1 2l(F^{-1}(q), x, y)dq,
\end{equation}
For the five datasets, we report the MAE and CRPS averaged over five trials.

\begin{table*}[!t]
\renewcommand{\arraystretch}{1.0}
\caption{Performance comparison of proposed method and two SOTA generative models at 70\% missing rate.}
    \label{tab:missing_70}
    \centering
    \begin{tabular}{ccc|cc|cc}
\toprule    
  &  \multicolumn{2}{c}{solar}  & \multicolumn{2}{c}{traffic} & \multicolumn{2}{c}{healthcare}\\ 
  Model  & MAE & CRPS & MAE & CRPS & MAE & CRPS \\
\midrule 
CSDI & 1.137  & 0.187  & \underline{3.382}  & \underline{0.083} & \underline{0.372} & \underline{0.406} \\
CSBI & \underline{1.105}  &  \underline{0.178} &  3.389 & 0.085 & 0.378 & 0.413\\
\textbf{Proposed method} & \textbf{0.668}  & \textbf{0.101}  &  \textbf{3.013} & \textbf{0.071} & \textbf{0.349} & \textbf{0.404} \\
\bottomrule 
\end{tabular}
\end{table*}

\begin{table*}[!t]
\renewcommand{\arraystretch}{1.0}
\caption{Performance comparison of proposed method and two SOTA generative models at 90\% missing rate.}
    \label{tab:missing_90}
    \centering
    \begin{tabular}{ccc|cc|cc}
\toprule    
  &  \multicolumn{2}{c}{solar}  & \multicolumn{2}{c}{traffic} & \multicolumn{2}{c}{healthcare}\\ 
  Model  & MAE & CRPS & MAE & CRPS & MAE & CRPS \\
\midrule 
CSDI & \underline{1.224}  & 0.224  & \underline{3.636}  & \underline{0.092} & \underline{0.484} & \underline{0.528} \\
CSBI & 1.228  &  \underline{0.213} &  3.685 & 0.095 & 0.498 & 0.577\\
\textbf{Proposed method}  & \textbf{0.766}  & \textbf{0.112}  &  \textbf{3.228} & \textbf{0.075} & \textbf{0.372} & \textbf{0.441} \\
\bottomrule 
\end{tabular}
\end{table*}

We compare our method with baseline models on five datasets that exhibit various missing patterns. To evaluate the performance under typical missing conditions, we artificially remove 50\% of the training data and using this data as ground truth. We report the results of MAE and CRPS for five trials in Table~\ref{tab:missing_50}. It is clear that our method consistently yields the best imputation results across both CRPS and MAE metrics in most cases. These findings highlight its strength not only in capturing temporal and feature dependencies for accurate deterministic imputations, but also in delivering high-quality probabilistic estimates that closely match the distribution of observations.

To further assess the baseline methods under higher missing rates, we compare our method with two state-of-the-art generative models (CSDI and CSBI) on three datasets at 70\% and 90\% missing rates, as shown in Table~\ref{tab:missing_70} and Table~\ref{tab:missing_90}. The results demonstrate that the proposed method surpasses both models in all experimental setups. These results emphasize that our method delivers exceptional imputation performance even when the available time series data is highly sparse, a challenging scenario often encountered in real-world applications. The performance gains are attributed to the innovative incorporation of quantile regression and sub-ensembles in our approach, which enables more accurate and reliable imputations compared to complex deep generative models.

\subsection{Ablation studies}

To assess the performance with varying ensemble member counts, we also set two additional quantile levels: $Q_2 = [0.1, 0.2,...,0.9], Q_3 = [0.05, 0.10,...,0.95]$. Similarly, we train our model at 50/70/90\% missing values for comparison in detail.

\begin{figure*}[t!]
    \centering
    \begin{subfigure}[b]{0.3\textwidth}
        \centering
        \includegraphics[width=\linewidth]{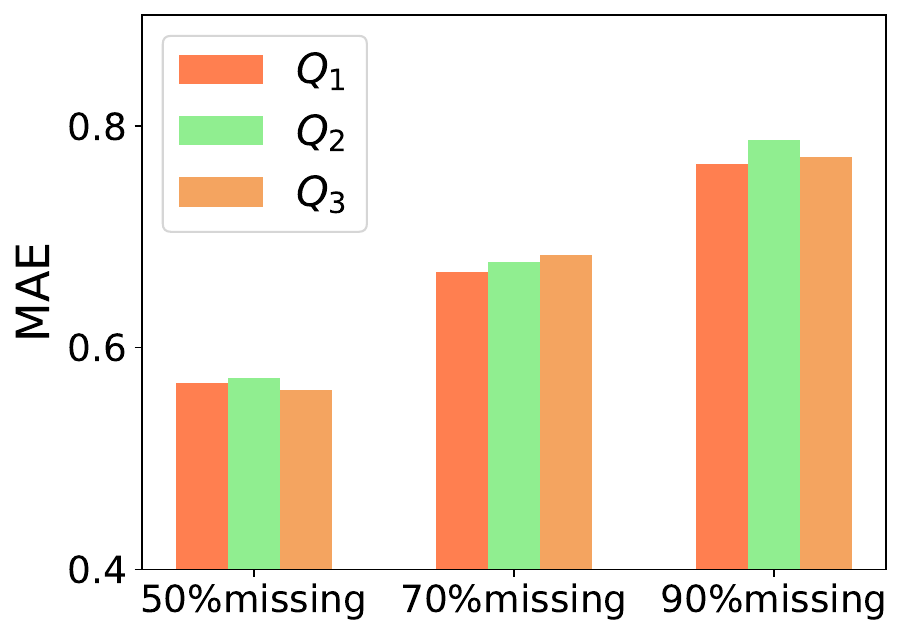}
        \caption{MAE of solar}
        \label{fig:solar_mae}
    \end{subfigure}
    \hfill
    \hspace{0.4cm}
    \begin{subfigure}[b]{0.3\textwidth}
        \centering
        \includegraphics[width=\linewidth]{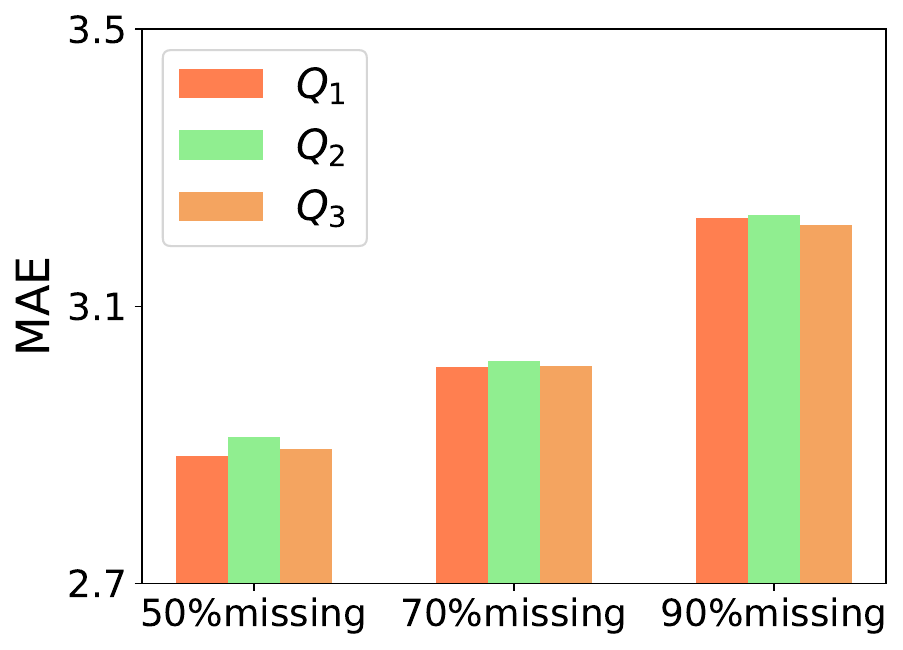}
        \caption{MAE of traffic}
        \label{fig:traffic_mae}
    \end{subfigure}
    \hfill
    \hspace{0.4cm}
    \begin{subfigure}[b]{0.3\textwidth}
        \centering
        \includegraphics[width=\linewidth]{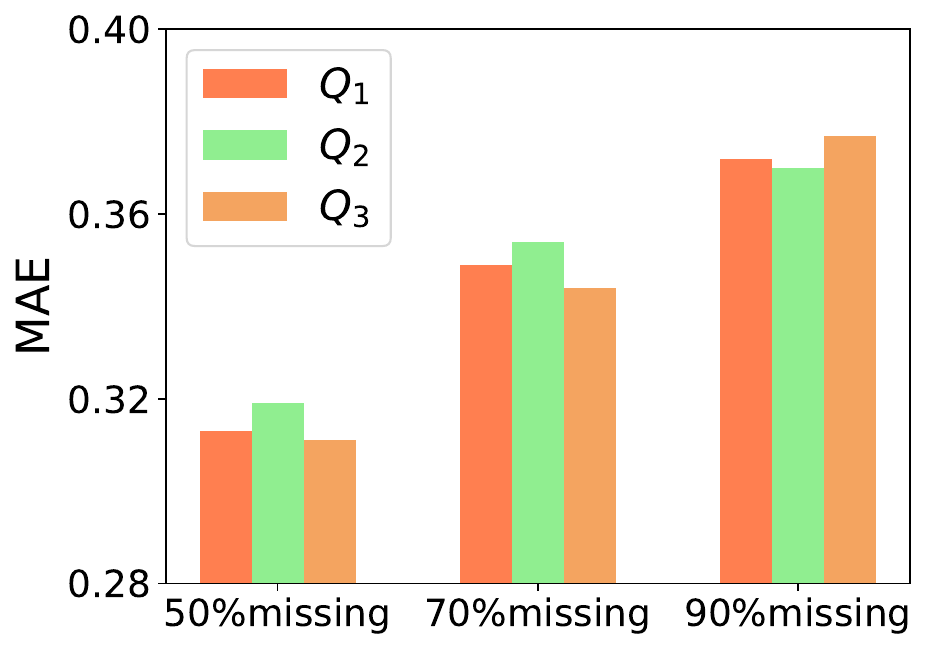}
        \caption{MAE of healthcare}
        \label{fig:healthcare_mae}
    \end{subfigure}
    \hfill
    \vspace{0.3cm}
    \begin{subfigure}[b]{0.3\textwidth}
        \centering
        \includegraphics[width=\linewidth]{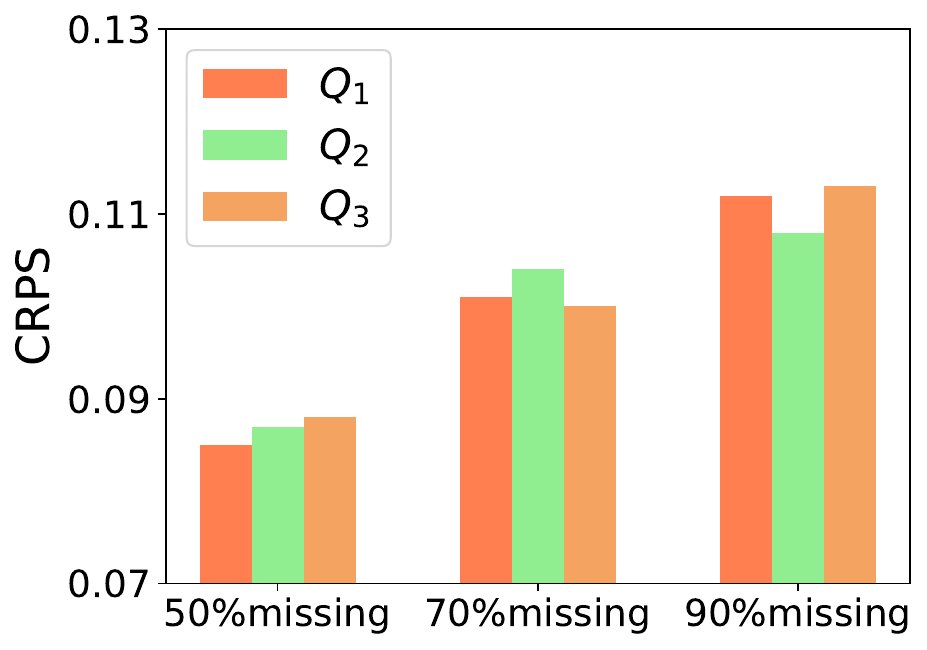}
        \caption{CRPS of solar}
        \label{fig:solar_crps}
    \end{subfigure}
    \hfill
    \begin{subfigure}[b]{0.3\textwidth}
        \centering
        \includegraphics[width=\linewidth]{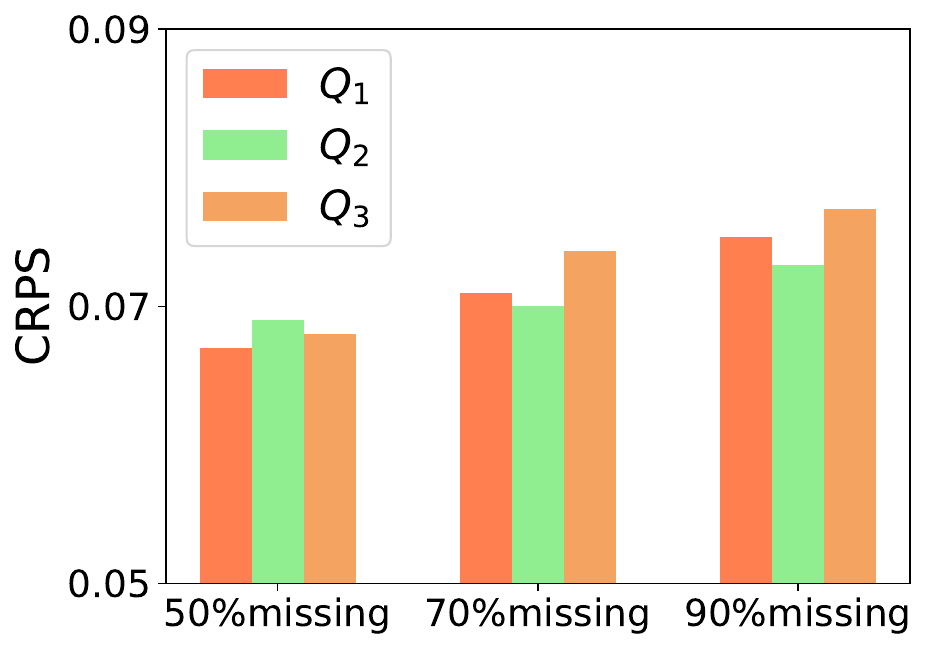}
        \caption{CRPS of traffic}
        \label{fig:traffic_crps}
    \end{subfigure}
    \hfill
    \begin{subfigure}[b]{0.3\textwidth}
        \centering
        \includegraphics[width=\linewidth]{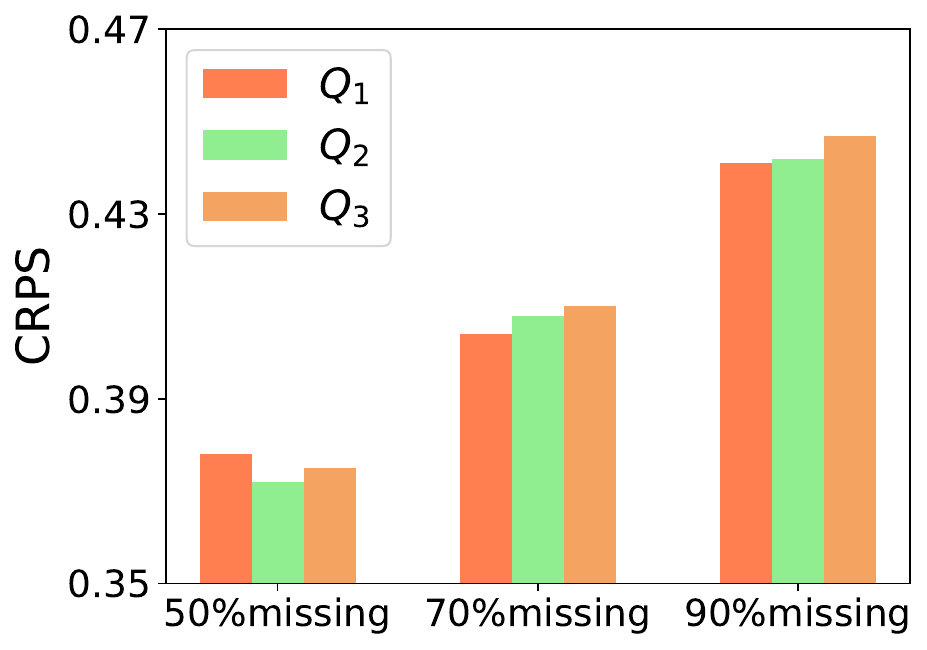}
        \caption{CRPS of healthcare}
        \label{fig:healthcare_crps}
    \end{subfigure}
    \caption{Performance of CRPS and MAE for three sets of quantile levels}
    \label{fig:quantiles}
\end{figure*}

Fig.~\ref{fig:quantiles} exhibits the performance of $Q_1, Q_2$ and $Q_3$ on three datasets. Through analysis, it can be concluded that $Q_1$ with five quantile levels is sufficient to achieve the best performance whether MAE or CRPS, verifying the robustness of our method for the number of ensemble members. 

\begin{table*}[t!]
\renewcommand{\arraystretch}{1.0}
\setlength{\tabcolsep}{3pt}
\caption{Performance of MAE and CRPS for Deep Ensembles and Quantile Sub-Ensembles with quantile levels $Q_1$ on three dataset at 50/70/90\% missing rate. }
    \label{tab:distinct ensembles}
    \centering
    \begin{tabular}{cccc|cc|cc}
\toprule    
  & & \multicolumn{2}{c}{solar}  & \multicolumn{2}{c}{traffic} & \multicolumn{2}{c}{healthcare}\\ 
  Missing & Model & MAE & CRPS & MAE & CRPS &  MAE & CRPS  \\
\midrule 
\multirow{2}{*}{50\%} & Deep Ensembles & 0.807 & 0.102 & 3.124 & 0.082 & 0.339 & 0.397 \\
& \textbf{Quantile Sub-Ensembles}  & \textbf{0.568}  & \textbf{0.085}  &  \textbf{2.884} & \textbf{0.067} & \textbf{0.313} & \textbf{0.378} \\
\midrule 
\multirow{2}{*}{70\%} & Deep Ensembles & 0.873 & 0.118 & 3.221 & 0.089 & 0.363 & 0.426 \\
& \textbf{Quantile Sub-Ensembles} & \textbf{0.668}  & \textbf{0.101}  &  \textbf{3.013} & \textbf{0.071} & \textbf{0.349} & \textbf{0.404} \\
\midrule 
\multirow{2}{*}{90\%} & Deep Ensembles & 0.913 & 0.127 & 3.388 & 0.093 & 0.391 & 0.456 \\
& \textbf{Quantile Sub-Ensembles}  & \textbf{0.766}  & \textbf{0.112}  &  \textbf{3.228} & \textbf{0.075} & \textbf{0.372} & \textbf{0.441} \\
\bottomrule 
\end{tabular}
\end{table*}

\begin{table*}[t!]
\renewcommand{\arraystretch}{1.0}
\setlength{\tabcolsep}{8pt}
\caption{Performance comparison of MAE and CRPS for imputation about the inclusion and exclusion of $\gamma$ of proposed method on three dataset at 50\% missing rate. }
    \label{tab:gamma}
    \centering
    \begin{tabular}{ccccccc}
\toprule    
  &  \multicolumn{2}{c}{solar}  & \multicolumn{2}{c}{traffic} & \multicolumn{2}{c}{healthcare}\\ 
  & MAE & CRPS & MAE & CRPS &  MAE & CRPS  \\
\midrule 
Exclusion ($\gamma$) & 0.637 & 0.094 & 2.992 & 0.078 & 0.328 & 0.391 \\
\textbf{Inclusion ($\gamma$)}  & \textbf{0.568}  & \textbf{0.085}  &  \textbf{2.884} & \textbf{0.067} & \textbf{0.313} & \textbf{0.378} \\
\bottomrule 
\end{tabular}
\end{table*}

\begin{figure*}[t!]
    \centering
    \begin{subfigure}[b]{0.3\textwidth}
        \centering
        \includegraphics[width=\linewidth]{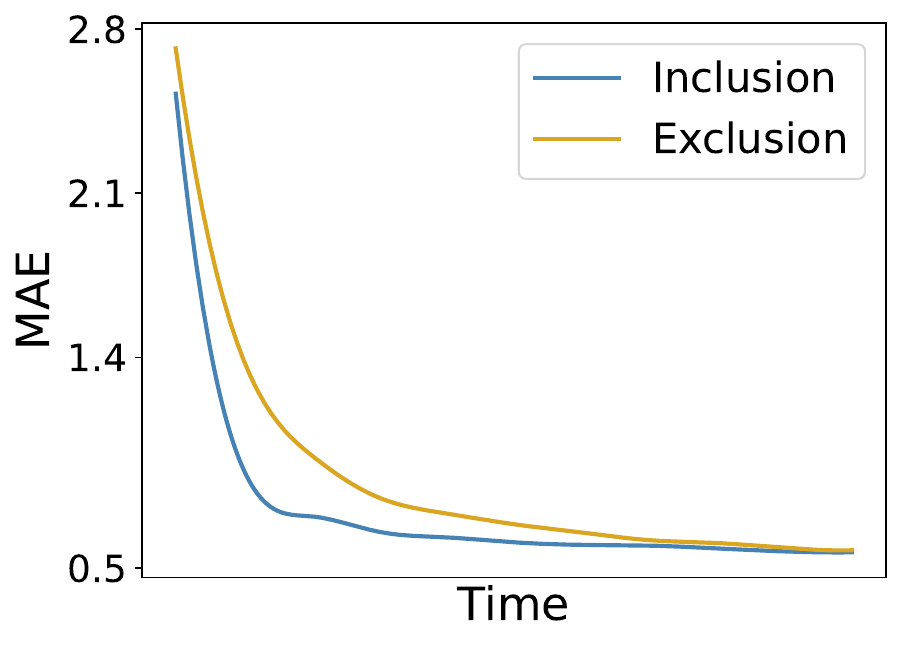}
        \caption{MAE of solar}
        \label{fig:loss_solar_mae}
    \end{subfigure}
    \hfill
    \hspace{0.4cm}
    \begin{subfigure}[b]{0.3\textwidth}
        \centering
        \includegraphics[width=\linewidth]{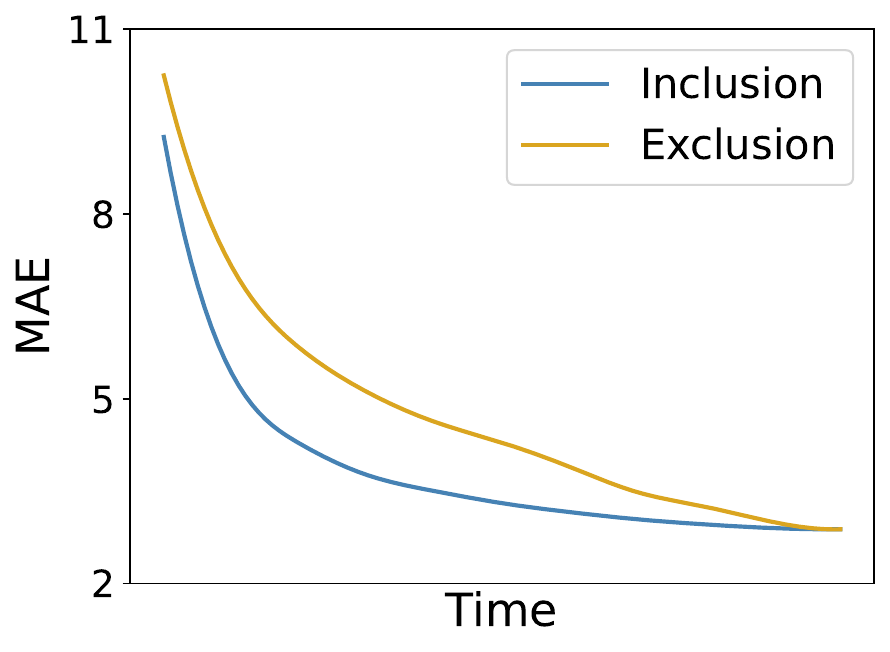}
        \caption{MAE of traffic}
        \label{fig:loss_traffic_mae}
    \end{subfigure}
    \hfill
    \hspace{0.4cm}
    \begin{subfigure}[b]{0.3\textwidth}
        \centering
        \includegraphics[width=\linewidth]{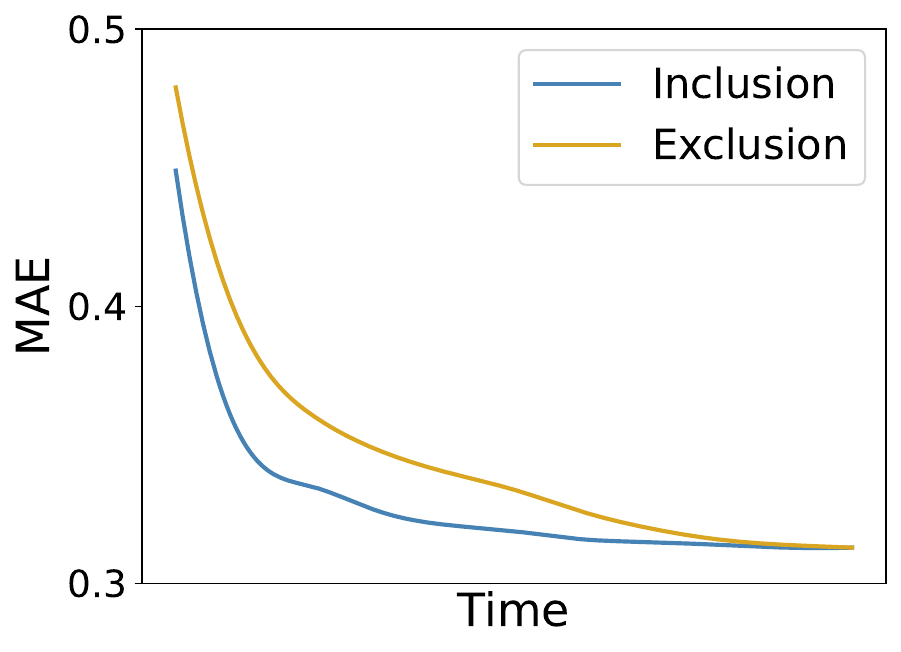}
        \caption{MAE of healthcare}
        \label{fig:loss_healthcare_mae}
    \end{subfigure}
    \hfill
    \vspace{0.3cm}
    \begin{subfigure}[b]{0.3\textwidth}
        \centering
        \includegraphics[width=\linewidth]{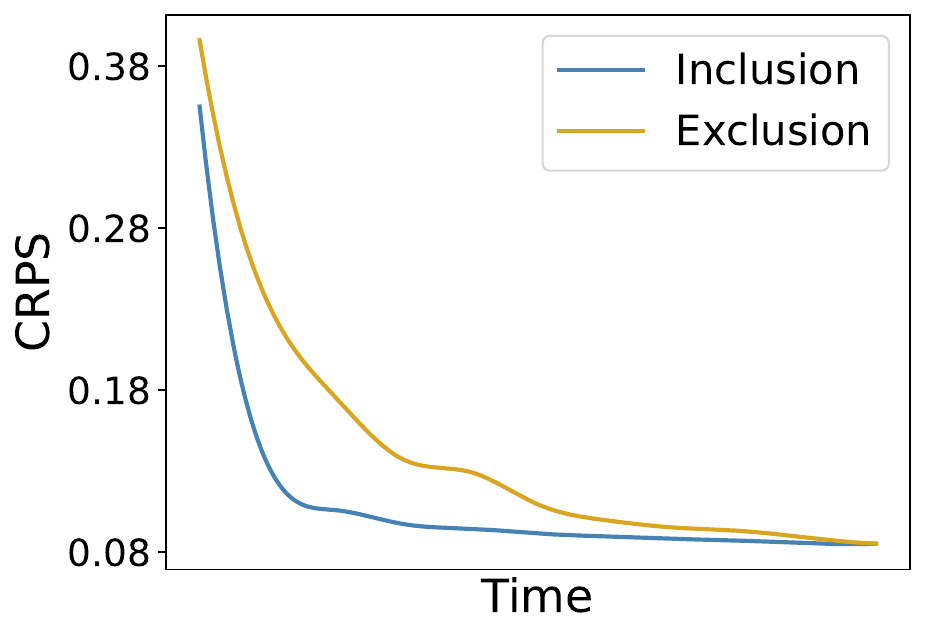}
        \caption{CRPS of solar}
        \label{fig:loss_solar_crps}
    \end{subfigure}
    \hfill
    \begin{subfigure}[b]{0.3\textwidth}
        \centering
        \includegraphics[width=\linewidth]{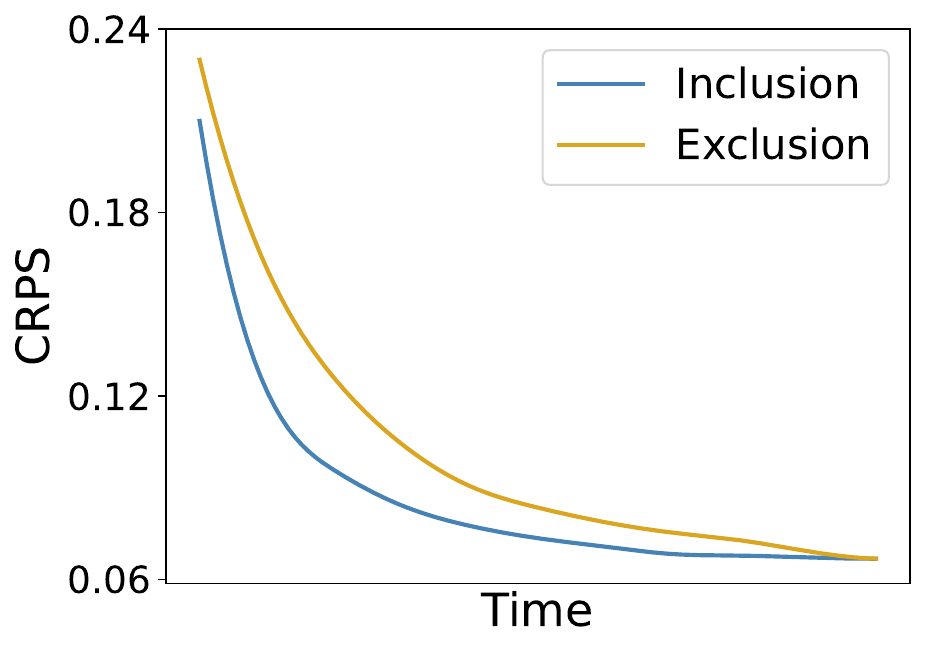}
        \caption{CRPS of traffic}
        \label{fig:loss_traffic_crps}
    \end{subfigure}
    \hfill
    \begin{subfigure}[b]{0.3\textwidth}
        \centering
        \includegraphics[width=\linewidth]{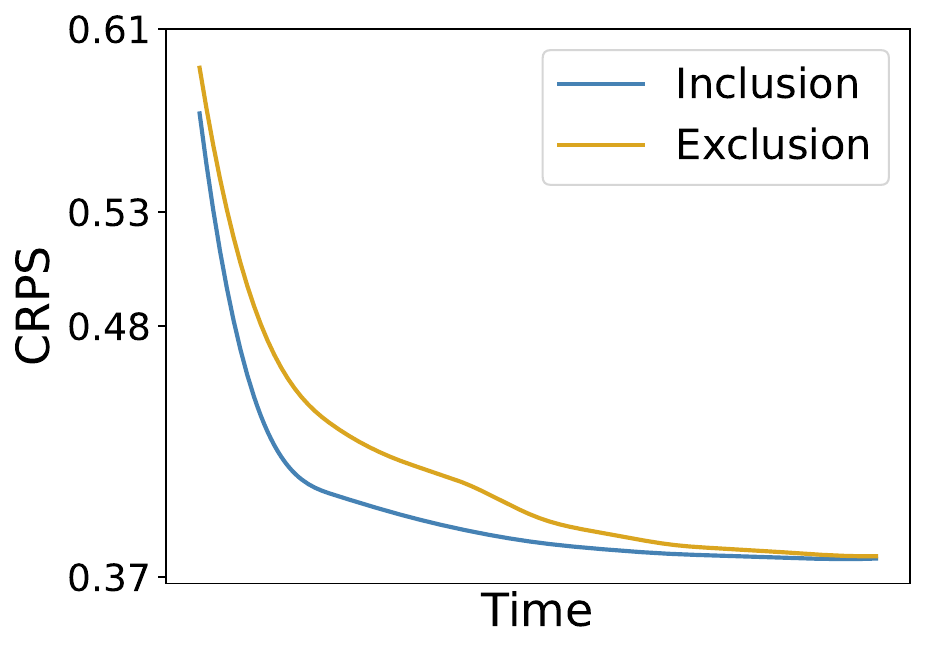}
        \caption{CRPS of healthcare}
        \label{fig:loss_healthcare_crps}
    \end{subfigure}
    \caption{Performance of MAE and CRPS for the model about the inclusion and exclusion of L1 and L2 losses on three dataset at 50\% missing rate.}
    \label{fig:loss}
\end{figure*}

We further compare Quantile Sub-Ensembles with classical Deep Ensembles as shown in Table~\ref{tab:distinct ensembles}. Overall it shows that Quantile Sub-Ensembles outperforms Deep Ensembles in all cases. This outcome is anticipated, as creating an ensemble with fewer layers than the entire model serves as a simplified version of the full ensemble. Integrating quantile regression this approach plays a pivotal role in reducing computational costs and estimating uncertainty quality.

To demonstrate the rationality of the structured choices made by the model, we have conducted experiments on whether the model structure includes the temporal decay factor $\gamma$. As shown in Table~\ref{tab:gamma}, removing $\gamma$ from the model leads to an increase in both MAE and CRPS of the imputed results. Therefore, it can be inferred that incorporating the temporal decay factor in the model effectively preserves the sequential temporal dependency information and enhances the model performance.

In Algorithm~\ref{algorithm_2}, we employ the optimization of weighted sums of L1 and L2 losses. To elucidate their role in model training and optimization, we conducted ablation experiments to assess the impact of including L1 and L2 losses in the model. The outcomes are illustrated in Fig.~\ref{fig:loss}. Upon meticulous analysis, it becomes apparent that the integration of L1 and L2 losses significantly enhances the training speed of the model, resulting in a more rapid convergence of results.

\begin{table*}[t!]
\renewcommand{\arraystretch}{1.0}
\setlength{\tabcolsep}{10pt}
\caption{The comparison of training time and inference time on healthcare dataset at 50\% missing rate}
    \label{tab:time}
    \centering
    \begin{tabular}{ccc}
\toprule    
 &  training time (s)  & inference time (s)  \\

\midrule 

CSDI & 8425.2 $\pm$ 163.4 & 4226.6$\pm$ 124.3 \\
Deep Ensembles & 1968.5 $\pm$ 26.3 &  13.2$\pm$ 1.5 \\
\textbf{Quantile Sub-Ensembles} & \textbf{621.7 $\pm$ 10.3} &  \textbf{2.8 $\pm$ 0.3} \\

\bottomrule 
\end{tabular}
\end{table*}

To evaluate the computational efficiency of the proposed method and measure the extent of its lead in training and inference speed, we conduct experiments on healthcare dataset at 50\% missing rate with same environment. The term ``training time" refers to the time it takes for a model to go from the initial training phase to complete fitting, while ``inference time" pertains to the time it takes for a trained model to perform data testing and inference. As shown in Table~\ref{tab:time}, the proposed method is significantly more computationally efficient than CSDI and Deep Ensembles, consuming much less training and inference time. Especially, CSDI needs a large amount of inference time due to performing multiple sampling and complex denoising process, which hinders deployment in real-world applications. Similarly, Deep Ensembles requires training multiple independent models from scratch, leading to a substantial increase in computational cost. In contrast, the proposed method achieves excellent performance while reducing computation, which is beneficial to practical deployment in real-time systems.

\begin{figure*}[t!]
    \centering
    \begin{subfigure}[b]{0.32\textwidth}
        \centering
        \includegraphics[width=\linewidth]{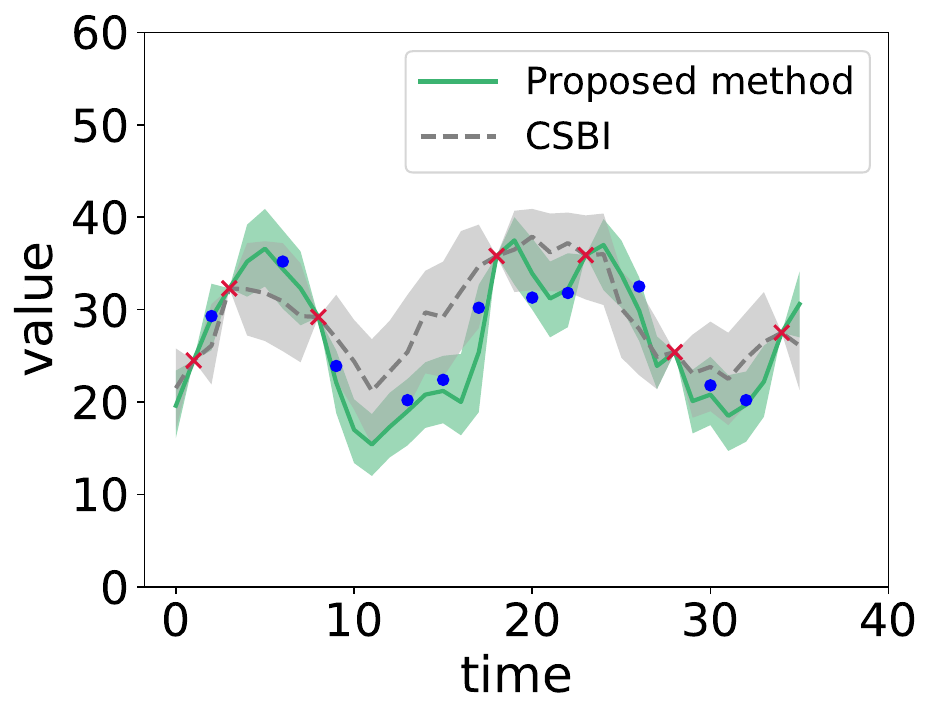}
        \caption{solar}
        \label{fig:example_solar}
    \end{subfigure}
    \begin{subfigure}[b]{0.32\textwidth}
        \centering
        \includegraphics[width=\linewidth]{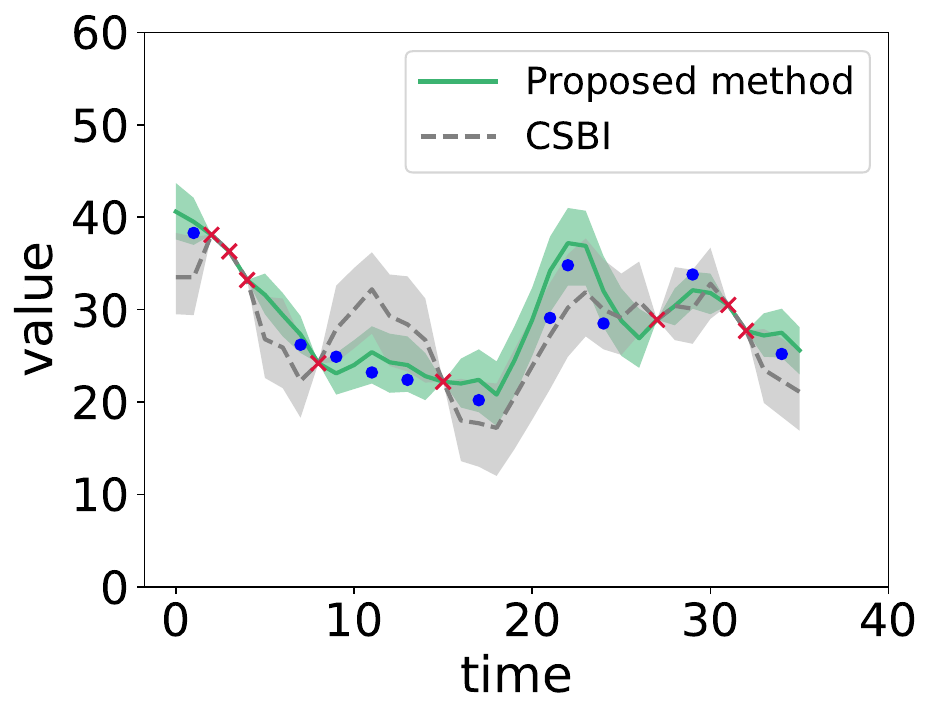}
        \caption{traffic}
        \label{fig:example_traffic}
    \end{subfigure}
    \begin{subfigure}[b]{0.32\textwidth}
        \centering
        \includegraphics[width=\linewidth]{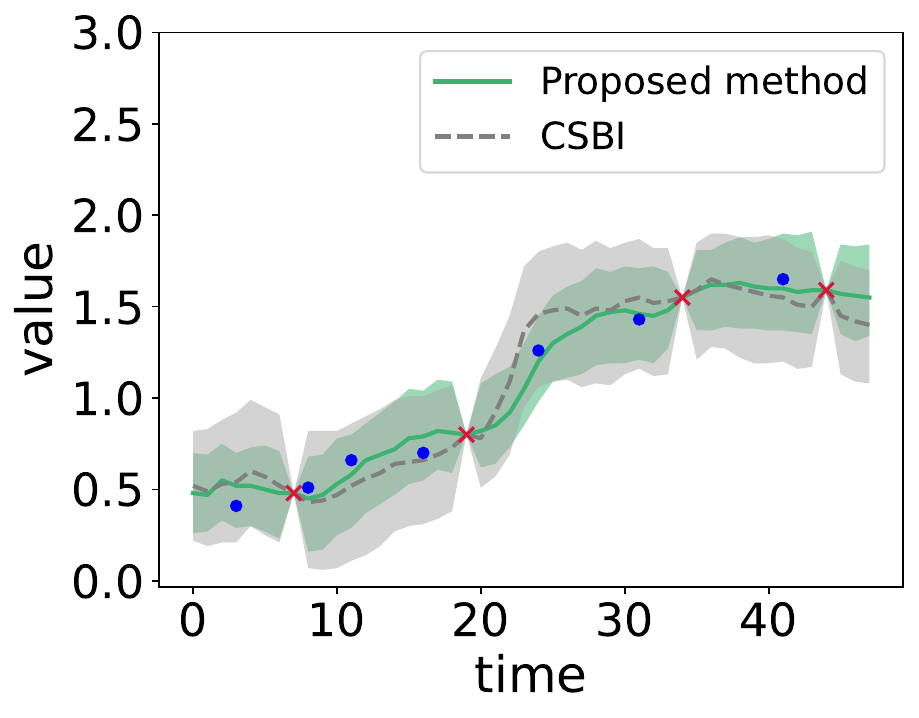}
        \caption{healthcare}
        \label{fig:example_health}
    \end{subfigure}
    \caption{A case study of time series imputation for the three datasets at 50\% missing rate. The red crosses represent the observed values, while the blue circles represent the ground truth. The lines represent the median values of the imputations and 5\% and 95\% quantiles are shown as the shade.}
    \label{fig:example}
\end{figure*}

\subsection{Case Study}
To provide an intuitional comparison, we give a case study to compare the proposed method with CSBI across three datasets, as shown in Figure~\ref{fig:example}. The proposed method (green solid line) exhibits better fitting to the ground truth, learns the temporal dependency better and provides more accurate imputations compared with CSBI (gray dashed line).  The area of the proposed method and CSBI between 5\% and 95\% quantiles are represented by the green shade and gray shade. The green shade is smaller than the gray shade, which indicates that the proposed method provides more reliable and reasonable imputation results.

\section{Conclusion}
In this paper, we introduced Quantile Sub-Ensembles, a novel uncertainty-aware approach for time series imputation that combines quantile-regression-based task networks. Our method addresses the critical issue of overconfidence in deep learning-based imputations by providing robust uncertainty estimates while maintaining high imputation accuracy, even in scenarios with a high missing rate. The ensemble framework enhances the predictive capabilities of the underlying model, offering a more reliable solution than conventional methods.

Through extensive experiments on five real-world datasets, our proposed method has been demonstrated its superiority over baseline approaches, particularly at higher missing rates. In comparison to state-of-the-art diffusion models like CSDI, Quantile Sub-Ensembles achieved comparable or better performance while significantly reducing computational overhead. This makes our approach not only effective but also efficient, enabling faster training and inference times.

In summary, our work provides an important step forward in the field of time series imputation by combining accuracy, uncertainty quantification, and computational efficiency, making it a valuable tool for real-world applications where data reliability and computational resources are critical. Future work could extend this framework to more complex datasets and further explore the use of uncertainty in decision-making processes.

\section*{Acknowledgements}
This paper is supported by 
the National Science Foundation of China Project (No. 62306098), 
Fundamental Research Funds for the Central Universities (No. JZ2024HGTB0256) and
the Open Project of Anhui Provincial Key Laboratory of Multimodal Cognitive Computation, Anhui University (No. MMC202412).


\section*{Clinical trial number}
Not applicable
\bibliographystyle{./bst/sn-mathphys-num}
\bibliography{sn-bibliography}

\end{document}